\documentclass[11pt,a4paper]{article}
\pdfoutput=1
\usepackage[hyperref]{naaclhlt2018}
\usepackage{times}
\usepackage{latexsym}
\usepackage{url}
\usepackage{graphicx}
\usepackage{subcaption}
\usepackage{verbatim}

\aclfinalcopy 

\title{A Hierarchical Latent Structure for Variational Conversation Modeling}

\author{Yookoon Park 
	\And
	Jaemin Cho \\
	Department of Computer Science and Engineering \& Center for Superintelligence\\
	Seoul National University, Korea\\
	{\tt \small yookoonpark@vision.snu.ac.kr}, {\tt \small \{jaemin895,gunhee\}@snu.ac.kr}\\	
	{\small \url{http://vision.snu.ac.kr/projects/vhcr}}\\
	\\\And
	Gunhee Kim \\}

\date{}

\usepackage{amsmath,amsfonts}
\usepackage{amssymb,amsopn}
\usepackage{bm} 
\usepackage{bbm}

\newcommand{\vct}[1]{\boldsymbol{\mathbf{#1}}} 
\newcommand{\mat}[1]{\boldsymbol{\mathbf{#1}}} 

\newlength{\widebarargwidth}
\newlength{\widebarargheight}
\newlength{\widebarargdepth}

\newcommand{\eat}[1]{}

\DeclareRobustCommand\onedot{\futurelet\@let@token\@onedot}
\def\onedot{. } 
\def\eg{\emph{e.g}\onedot} 
\def\ie{\emph{i.e}\onedot}

\begin{document}

\maketitle

\begin{abstract}
Variational autoencoders (VAE) combined with hierarchical RNNs have emerged as a powerful framework for conversation modeling. However, they suffer from the notorious degeneration problem, where the decoders learn to ignore latent variables and reduce to vanilla RNNs. 
We empirically show that this degeneracy occurs mostly due to two reasons.
First, the expressive power of hierarchical RNN decoders is often high enough to model the data using only its decoding distributions without relying on the latent variables. 
Second, the conditional VAE structure whose generation process is conditioned on a context, makes the range of training targets very sparse; that is, the RNN decoders can easily overfit to the training data ignoring the latent variables.
To solve the degeneration problem, we propose a novel model named \textit{Variational Hierarchical Conversation RNNs} (VHCR), involving two key ideas of (1) using a hierarchical structure of latent variables, and (2) exploiting an \textit{utterance drop} regularization. 
With evaluations on two datasets of Cornell Movie Dialog and Ubuntu Dialog Corpus,
we show that our VHCR successfully utilizes latent variables and outperforms state-of-the-art models for conversation generation.
Moreover, it can perform several new utterance control tasks, thanks to its hierarchical latent structure. 
\end{abstract}

\section{Introduction}
\label{sec:intro}

Conversation modeling has been a long interest of natural language research. 
Recent approaches for data-driven conversation modeling mostly build upon recurrent neural networks (RNNs) \citep{neural_conversation, context_conversation, neural_respond, gan_dialogue, hrnn_dialogue}. \citet{hrnn_dialogue} use a hierarchical RNN structure to model the context of conversation. \citet{vhred} further exploit an utterance latent variable in the hierarchical RNNs by incorporating the variational autoencoder (VAE) framework \cite{vae, vae2}. 

VAEs enable us to train a latent variable model for natural language modeling, which grants us several advantages. 
First, latent variables can learn an interpretable holistic representation, such as topics, tones, or high-level syntactic properties. 
Second, latent variables can model inherently abundant variability of natural language by encoding its global and long-term structure, 
which is hard to be captured by shallow generative processes (\eg vanilla RNNs) where the only source of stochasticity comes from the sampling of output words. 

In spite of such appealing properties of latent variable models for natural language modeling,
VAEs suffer from the notorious \textit{degeneration} problem \citep{sentence_vae,lossy_vae} that occurs when a VAE is combined with a powerful decoder such as autoregressive RNNs.
This issue makes VAEs ignore latent variables, and eventually behave as vanilla RNNs. 
\citet{lossy_vae} also note this degeneration issue by showing that a VAE with a RNN decoder prefers to model the data using its decoding distribution rather than using latent variables, from bits-back coding perspective. 
To resolve this issue, several heuristics have been proposed to weaken the decoder, enforcing the models to use latent variables.
For example, \citet{sentence_vae} 
propose some heuristics, including \textit{KL annealing} and \textit{word drop} regularization. 
However, these heuristics cannot be a complete solution; for example, we observe that they fail to prevent the degeneracy in VHRED \citep{vhred}, a conditional VAE model equipped with hierarchical RNNs for conversation modeling. 

The objective of this work is to propose a novel VAE model that significantly alleviates the degeneration problem. 
Our analysis reveals that the causes of the degeneracy are two-fold.
First, the hierarchical structure of autoregressive RNNs is powerful enough to predict a sequence of utterances without the need of latent variables, even with the word drop regularization. 
Second, we newly discover that the conditional VAE structure where an utterance is generated conditioned on context, \ie a previous sequence of utterances, induces severe data sparsity.
Even with a large-scale training corpus, there only exist very few target utterances when conditioned on the context. 
Hence, the hierarchical RNNs can easily memorize the context-to-utterance relations without relying on latent variables. 

We propose a novel model named \textit{Variational Hierarchical Conversation RNN} (VHCR),
which involves two novel features to alleviate this problem.
First, we introduce a global conversational latent variable along with local utterance latent variables to build a hierarchical latent structure.
Second, we propose a new regularization technique called \textit{utterance drop}.
We show that our hierarchical latent structure is not only crucial for facilitating the use of latent variables in conversation modeling, but also delivers several additional advantages, including gaining control over the global context in which the conversation takes place. 

Our major contributions are as follows:

(1) We reveal that the existing conditional VAE model with hierarchical RNNs for conversation modeling (\eg \citep{vhred}) still suffers from the degeneration problem, 
and this problem is caused by data sparsity per context that arises from the conditional VAE structure, as well as the use of powerful hierarchical RNN decoders.

(2) We propose a novel variational hierarchical conversation RNN (VHCR), which has two distinctive features: a hierarchical latent structure and a new regularization of utterance drop.
To the best of our knowledge, our VHCR is the first VAE conversation model that exploits the hierarchical latent structure. 

(3) With evaluations on two benchmark datasets of Cornell Movie Dialog \citep{movie_dialogue} and Ubuntu Dialog Corpus \citep{ubuntu_corpus},
we show that our model improves the conversation performance in multiple metrics over state-of-the-art methods, including HRED \citep{hrnn_dialogue}, and VHRED \citep{vhred} with existing degeneracy solutions such as the word drop \citep{sentence_vae}, and the bag-of-words loss \citep{vae_bow}.

\section{Related Work}
\label{sec:related_work}

\textbf{Conversation Modeling}.
One popular approach for conversation modeling is to use RNN-based encoders and decoders, such as~\citep{neural_conversation, context_conversation, neural_respond}.
Hierarchical recurrent encoder-decoder (HRED) models~\citep{hrnn_query, hrnn_dialogue, vhred} 
consist of utterance encoder and decoder, and a context RNN
which runs over utterance representations to model long-term temporal structure of conversation.

Recently, latent variable models such as VAEs have been adopted in language modeling~\citep{sentence_vae, vnmt, vhred}. 
The VHRED model~\citep{vhred} integrates the VAE with the HRED to model Twitter and Ubuntu IRC conversations by introducing an utterance latent variable. This makes a conditional VAE where the generation process is conditioned on the context of conversation. 
\citet{vae_bow} further make use of discourse act labels to capture the diversity of conversations. 

\textbf{Degeneracy of Variational Autoencoders}.
For sequence modeling, VAEs are often merged with the RNN encoder-decoder structure~\citep{sentence_vae, vhred, vae_bow} where the encoder predicts the posterior distribution of a latent variable $\vct{z}$, and the decoder models the output distributions conditioned on $\vct{z}$. 
However, \citet{sentence_vae} report that a VAE with a RNN decoder easily degenerates; that is, it learns to ignore the latent variable $\vct{z}$ and falls back to a vanilla RNN. They propose two techniques to alleviate this issue: \textit{KL annealing} and \textit{word drop}. 
\citet{lossy_vae} interpret this degeneracy in the context of bits-back coding and show that a VAE equipped with autoregressive models such as RNNs often ignores the latent variable to minimize the code length needed for describing data. They propose to constrain the decoder to selectively encode the information of interest in the latent variable. However, their empirical results are limited to an image domain. 
\citet{vae_bow} use an auxiliary bag-of-words loss on the latent variable to force the model to use $\vct{z}$. That is, they train an auxiliary network that predicts bag-of-words representation of the target utterance based on $\vct{z}$. Yet this loss works in an opposite direction to the original objective of VAEs that minimizes the minimum description length. Thus, it may be in danger of forcibly moving the information that is better modeled in the decoder to the latent variable.

\section{Approach}
\label{sec:approach}
We assume that
the training set consists of $N$ i.i.d samples of conversations $\{\vct{c}_1, \vct{c}_2, ..., \vct{c}_N \}$ where each $\vct{c}_i$ is a sequence of utterances (\ie sentences) $\{\vct{x}_{i1}, \vct{x}_{i2}, ..., \vct{x}_{in_i}\}$. Our objective is to learn the parameters of a generative network $\vct{\theta}$ using Maximum Likelihood Estimation (MLE):
\begin{align}
\arg \max_{\vct{\theta}} \sum_i \log p_{\vct{\theta}}(\vct{c}_i) 
\end{align}

We first briefly review the VAE, and explain the degeneracy issue before presenting our model.

\subsection{Preliminary: Variational Autoencoder}
\label{sec:pre_vae}

We follow the notion of \citet{vae}. 
A datapoint $\vct{x}$ is generated from a latent variable $\vct{z}$, which is sampled from some prior distribution $p(\vct{z})$, typically a standard Gaussian distribution $\mathcal{N}(\vct{z} | \vct{0}, \mat{I})$. 
We assume parametric families for conditional distribution $p_{\vct{\theta}}(\vct{x}|\vct{z})$. 
Since it is intractable to compute the log-marginal likelihood $\log p_{\vct{\theta}}(\vct{x})$, we approximate the intractable true posterior $p_{\vct{\theta}}(\vct{z}|\vct{x})$ with a recognition model $q_{\vct{\phi}}(\vct{z}|\vct{x})$ to maximize the \textit{variational lower-bound}:
\begin{align}\
\label{elbo1}
&\log p_{\vct{\theta}}(\vct{x}) \ge \mathcal{L}(\vct{\theta}, \vct{\phi}; \vct{x})\\
&= \mathbb{E}_{q_{\vct{\phi}}(\vct{z}|\vct{x})}[-\log q_{\vct{\phi}}(\vct{z}|\vct{x}) + \log p_{\vct{\theta}}(\vct{x}, \vct{z})] \nonumber \\
&= -D_{KL}(q_{\vct{\phi}}(\vct{z}|\vct{x}) \| p(\vct{z})) \hspace{-1pt}+\hspace{-1pt} \mathbb{E}_{q_{\vct{\phi}}(\vct{z}|\vct{x})}[\log p_{\vct{\theta}}(\vct{x}|\vct{z})]  \nonumber 
\end{align}
Eq.~\ref{elbo1} is decomposed into two terms: KL divergence term and reconstruction term. Here, KL divergence measures the amount of information encoded in the latent variable $\vct{z}$. In the extreme where KL divergence is zero, the model completely ignores $\vct{z}$, \textit{i.e.} it degenerates.
The expectation term can be stochastically approximated by sampling $\vct{z}$ from the variational posterior $q_{\vct{\phi}}(\vct{z}|\vct{x})$. The gradients to the recognition model can be efficiently estimated using the \textit{reparameterization} trick \citep{vae}.

\subsection{VHRED}
\label{sec:vhred}

\citet{vhred} propose Variational Hierarchical Recurrent Encoder Decoder~(VHRED) model for conversation modeling. 
It integrates an utterance latent variable $\vct{z}^{\text{utt}}_t$ into the HRED structure~\citep{hrnn_query} which consists of three RNN components: \textit{encoder RNN}, \textit{context RNN}, and \textit{decoder RNN}. Given a previous sequence of utterances $\vct{x}_1 , ... \vct{x}_{t-1}$ in a conversation, the VHRED generates the next utterance $\vct{x}_{t}$ as:
\begin{align}
\label{vhred_encoder} \vct{h}^{\text{enc}}_{t-1} &= f^{\text{enc}}_{\vct{\theta}}(\vct{x}_{t-1}) \\
\label{vhred_context} \vct{h}^{\text{cxt}}_t &= f^{\text{cxt}}_{\vct{\theta}}(\vct{h}^{\text{cxt}}_{t-1}, \vct{h}^{\text{enc}}_{t-1}) \\
\label{vhred_prior} p_{\vct{\theta}}(\vct{z}^{\text{utt}}_t | \vct{x}_{<t}) &= \mathcal{N}(\vct{z} | \vct{\mu}_t, \vct{\sigma}_t \vct{I}) \\
\mbox{where } \vct{\mu}_t &= \text{MLP}_{\vct{\theta}}(\vct{h}^{\text{cxt}}_t) \\
\label{vhred_prior_var}\vct{\sigma}_t &= \text{Softplus(MLP}_{\vct{\theta}}(\vct{h}^{\text{cxt}}_t)) \\
\label{vhred_decoder} p_{\vct{\theta}}(\vct{x}_t | \vct{x}_{<t}) &= f^{\text{dec}}_{\vct{\theta}}(\vct{x} |\vct{h}^{\text{cxt}}_t, \vct{z}^{\text{utt}}_t)
\end{align}
At time step $t$, the encoder RNN $f^{\text{enc}}_{\vct{\theta}}$ takes the previous utterance $\vct{x}_{t-1}$ and produces an encoder vector $\vct{h}^{\text{enc}}_{t-1}$ (Eq.~\ref{vhred_encoder}).
The context RNN $f^{\text{cxt}}_{\vct{\theta}}$ models the context of the conversation by updating its hidden states using the encoder vector (Eq.~\ref{vhred_context}).
The context $\vct{h}^{\text{cxt}}_t$ defines the conditional prior $p_{\vct{\theta}}(\vct{z}^{\text{utt}}_t | \vct{x}_{<t})$, which is a factorized Gaussian distribution whose mean $ \vct{\mu}_t$ and diagonal variance $\vct{\sigma}_t$ are given by feed-forward neural networks (Eq.~\ref{vhred_prior}-\ref{vhred_prior_var}). 
Finally the decoder RNN $f^{\text{dec}}_{\vct{\theta}}$ generates the utterance $\vct{x}_{t}$, conditioned on the context vector $\vct{h}^{\text{cxt}}_t$ and the latent variable $\vct{z}^{\text{utt}}_t$ (Eq.~\ref{vhred_decoder}). 
We make two important notes: (1) the context RNN can be viewed as a high-level decoder, and together with the decoder RNN, they comprise a hierarchical RNN decoder. (2) VHRED follows a conditional VAE structure where each utterance $\vct{x}_{t}$ is generated conditioned on the context $\vct{h}^{\text{cxt}}_t$ (Eq.~\ref{vhred_prior}-\ref{vhred_decoder}). 

The variational posterior is  a factorized Gaussian distribution where the mean and the diagonal variance are predicted from the target utterance and the context as follows:
\begin{align}
q_{\vct{\phi}}(\vct{z}^{\text{utt}}_t | \vct{x}_{\le t}) &= \mathcal{N}(\vct{z} | \vct{\mu}'_t, \vct{\sigma}'_t I)\\
\mbox{where } \vct{\mu}'_t &= \text{MLP}_{\vct{\phi}}(\vct{x}_t, \vct{h}^{\text{cxt}}_t) \\
\vct{\sigma}'_t &= \text{Softplus(MLP}_{\vct{\phi}}(\vct{x}_t, \vct{h}^{\text{cxt}}_t))
\end{align}

\begin{figure}[t]
	\centering
	\includegraphics[width=0.47\textwidth]{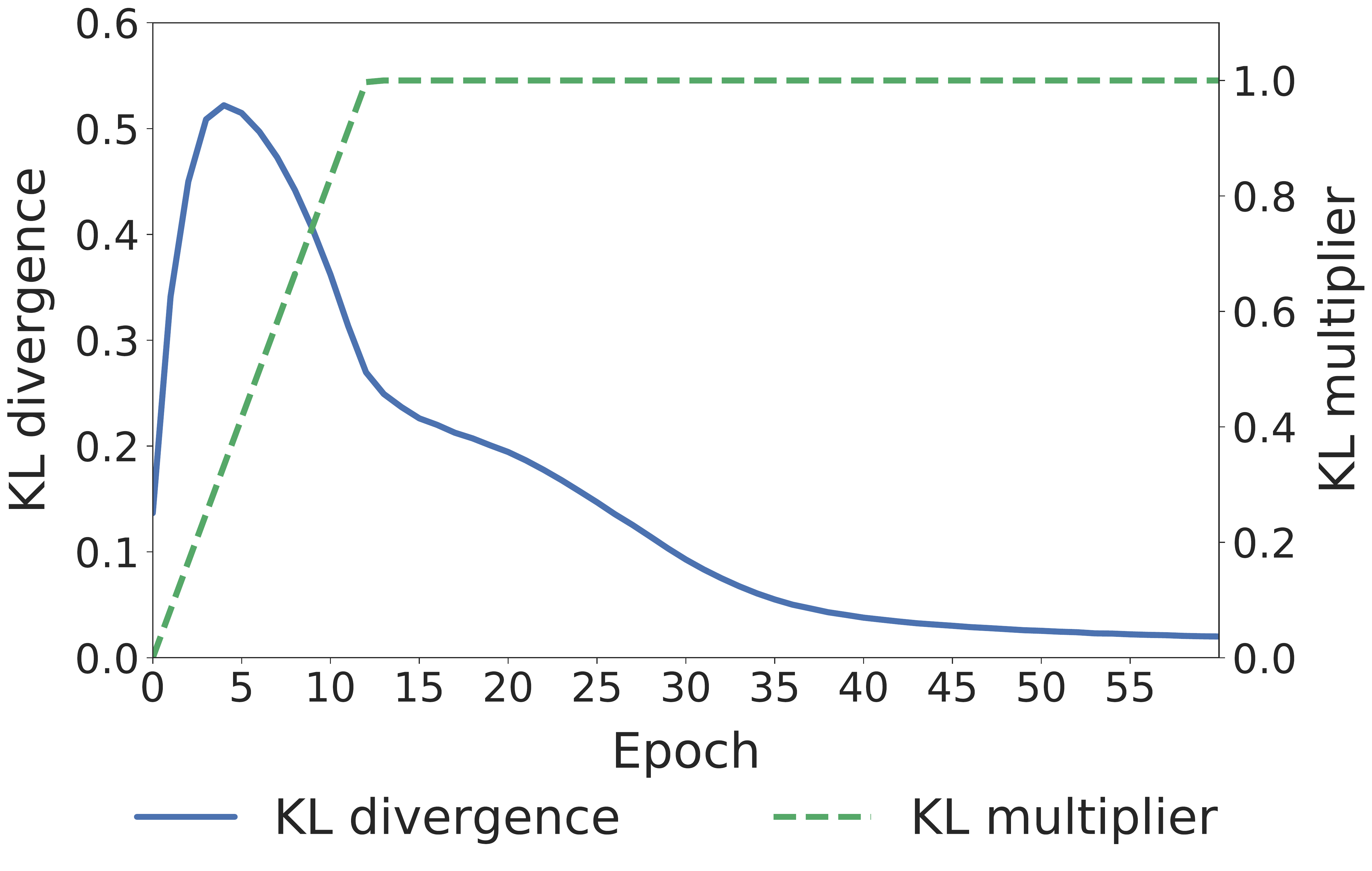}
	\vspace{-0.3cm}
	\caption{\textbf{Degeneration of VHRED}. The KL divergence term continuously decreases as training proceeds, meaning that the decoder ignores the latent variable $\vct{z}^{utt}$. We train the VHRED on Cornell Movie Dialog Corpus with word drop and KL annealing.}
	\label{fig:degeneration}
	\vspace{-0.3cm}
\end{figure}

\subsection{The Degeneration Problem}
\label{sec:degeneration}

A known problem of a VAE that incorporates an autoregressive RNN decoder is the degeneracy that ignores the latent variable $\vct{z}$.
In other words, the KL divergence term in Eq.~\ref{elbo1} goes to zero and the decoder fails to learn any dependency between the latent variable and the data. 
Eventually, the model behaves as a vanilla RNN. 
This problem is first reported in the sentence VAE~\citep{sentence_vae}, in which following two heuristics are proposed to alleviate the problem by weakening the decoder. 

First, the \textit{KL annealing} scales the KL divergence term of Eq.~\ref{elbo1} using a KL multiplier $\lambda$, which gradually increases from 0 to 1 during training:
\begin{align}
\mathcal{\tilde{L}}(\vct{\theta}, \vct{\phi}; \vct{x}) 
= -\lambda D_{KL}(q_{\vct{\phi}}(\vct{z}|\vct{x}) \| p(\vct{z})) \\
+ \mathbb{E}_{q_{\vct{\phi}}(\vct{z}|\vct{x})}[\log p_{\vct{\theta}}(\vct{x}|\vct{z})] \nonumber
\end{align}
This helps the optimization process to avoid local optima of zero KL divergence in early training.
Second, the \textit{word drop} regularization randomly replaces some conditioned-on word tokens in the RNN decoder with the generic unknown word token (UNK) during training.
Normally, the RNN decoder predicts each next word in an autoregressive manner, conditioned on the previous sequence of ground truth (GT) words. By randomly replacing a GT word with an UNK token, the word drop regularization weakens the autoregressive power of the decoder and forces it to rely on the latent variable to predict the next word. 
The word drop probability is normally set to 0.25, since using a higher probability may degrade the model performance~\citep{sentence_vae}. 

However, we observe that these tricks do not solve the degeneracy for the VHRED in conversation modeling.
An example in Fig.~\ref{fig:degeneration} shows that the VHRED learns to ignore the utterance latent variable as the KL divergence term falls to zero. 

\subsection{Empirical Observation on Degeneracy}

\begin{figure}[t]
	\centering
	\includegraphics[width=0.47\textwidth]{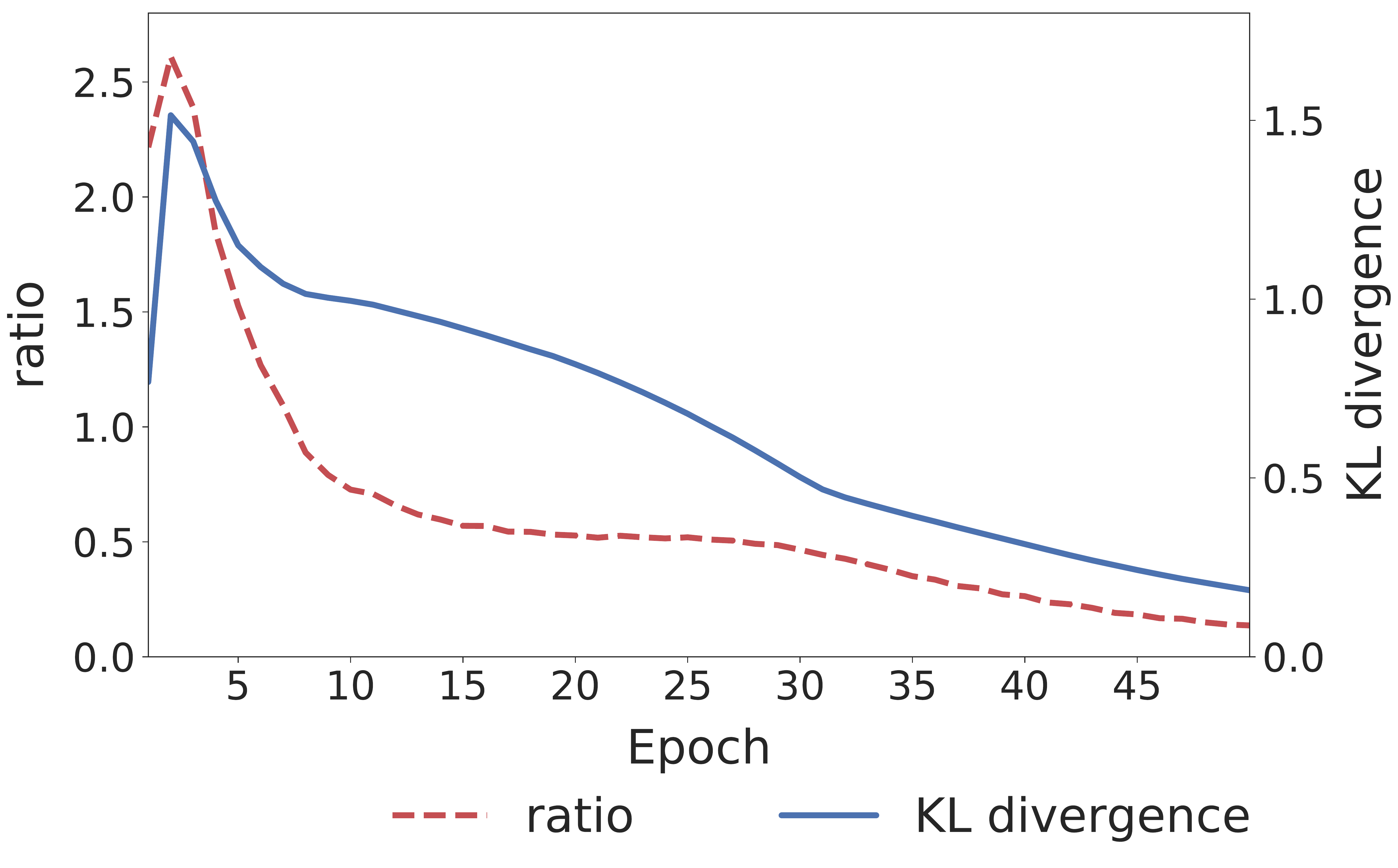}
	\vspace{-0.3cm}
	\caption{\textbf{The average ratio $\mathbb{E}[\vct{\sigma}^2_t] / \mbox{Var}(\vct{\mu}_t)$ when the decoder is only conditioned on $\vct{z}^{\text{utt}}_t$}. 
	The ratio drops to zero as training proceeds, indicating that the conditional priors $p_{\vct{\theta}}(\vct{z}^{\text{utt}}_t|\vct{x}_{<t})$ degenerate to separate point masses. 
	}
	\vspace{-0.3cm}
	\label{fig:var_ratio}
\end{figure}

The decoder RNN  of the VHRED in Eq.~\ref{vhred_decoder} conditions on two information sources: deterministic $\vct{h}^{\text{cxt}}_t$ and stochastic $\vct{z}^{\text{utt}}$. 
In order to check whether the presence of deterministic source $\vct{h}^{\text{cxt}}_t$ causes the degeneration,
we drop the deterministic $\vct{h}^{\text{cxt}}_t$ and condition the decoder only on the stochastic utterance latent variable $\vct{z}^{\text{utt}}$:

\begin{align}
\label{stochastic_decoder}
p_{\vct{\theta}}(\vct{x}_t | \vct{x}_{<t}) &= f^{\text{dec}}_{\vct{\theta}}(\vct{x} |\vct{z}^{\text{utt}}_t) 
\end{align}
While this model achieves higher values of KL divergence than original VHRED, as training proceeds it again degenerates with the KL divergence term reaching zero (Fig.~\ref{fig:var_ratio}). 

To gain an insight of the degeneracy, we examine how the conditional prior $p_{\vct{\theta}}(\vct{z}^{\text{utt}}_t|\vct{x}_{<t})$ (Eq.~\ref{vhred_prior}) of the utterance latent variable changes during training, using the model above (Eq. \ref{stochastic_decoder}). Fig.~\ref{fig:var_ratio} plots the ratios of $\mathbb{E}[\vct{\sigma}^2_t] / \mbox{Var}(\vct{\mu}_t)$, where $\mathbb{E}[\vct{\sigma}^2_t]$ indicates the \textit{within variance} of the priors, and $\mbox{Var}(\vct{\mu}_t)$ is the \textit{between variance} of the priors. 
Note that traditionally this ratio is closely related to \textit{Analysis of Variance (ANOVA)} \citep{anova}.
The ratio gradually falls to zero, implying that the priors degenerate to separate point masses as training proceeds. Moreover, we find that the degeneracy of priors coincide with the degeneracy of KL divergence, as shown in (Fig.~\ref{fig:var_ratio}). This is intuitively natural: if the prior is already narrow enough to specify the target utterance, there is little pressure to encode any more information in the variational posterior for reconstruction of the target utterance.

This empirical observation implies that the fundamental reason behind the degeneration may originate from combination of two factors: (1) strong expressive power of the hierarchical RNN decoder and (2) training data sparsity caused by the conditional VAE structure. 
The VHRED is trained to predict a next target utterance $\vct{x}_{t}$ conditioned on the context $\vct{h}^{\text{cxt}}_t$ which encodes information about previous utterances $\{\vct{x}_{1}, \dots, \vct{x}_{t-1}\}$. However, conditioning on the context makes the range of training target $\vct{x}_{t}$ very sparse; even in a large-scale conversation corpus such as Ubuntu Dialog \citep{ubuntu_corpus}, there exist one or very few target utterances per context. 
Therefore, hierarchical RNNs, given their autoregressive power, can easily overfit to training data without using the latent variable. 
Consequently, the VHRED will not encode any information in the latent variable, i.e. it degenerates.
It explains why the word drop fails to prevent the degeneracy in the VHRED.
The word drop only regularizes the decoder RNN; however, the context RNN is also powerful enough to predict a next utterance in a given context even with the weakened decoder RNN.  
Indeed we observe that using a larger word drop probability such as 0.5 or 0.75 only slows down, but fails to stop the KL divergence from vanishing.

\subsection{Variational Hierarchical Conversation RNN (VHCR)}
\label{sec:vhcr}

\begin{figure}[t]
	\centering
	\includegraphics[width=0.47\textwidth]{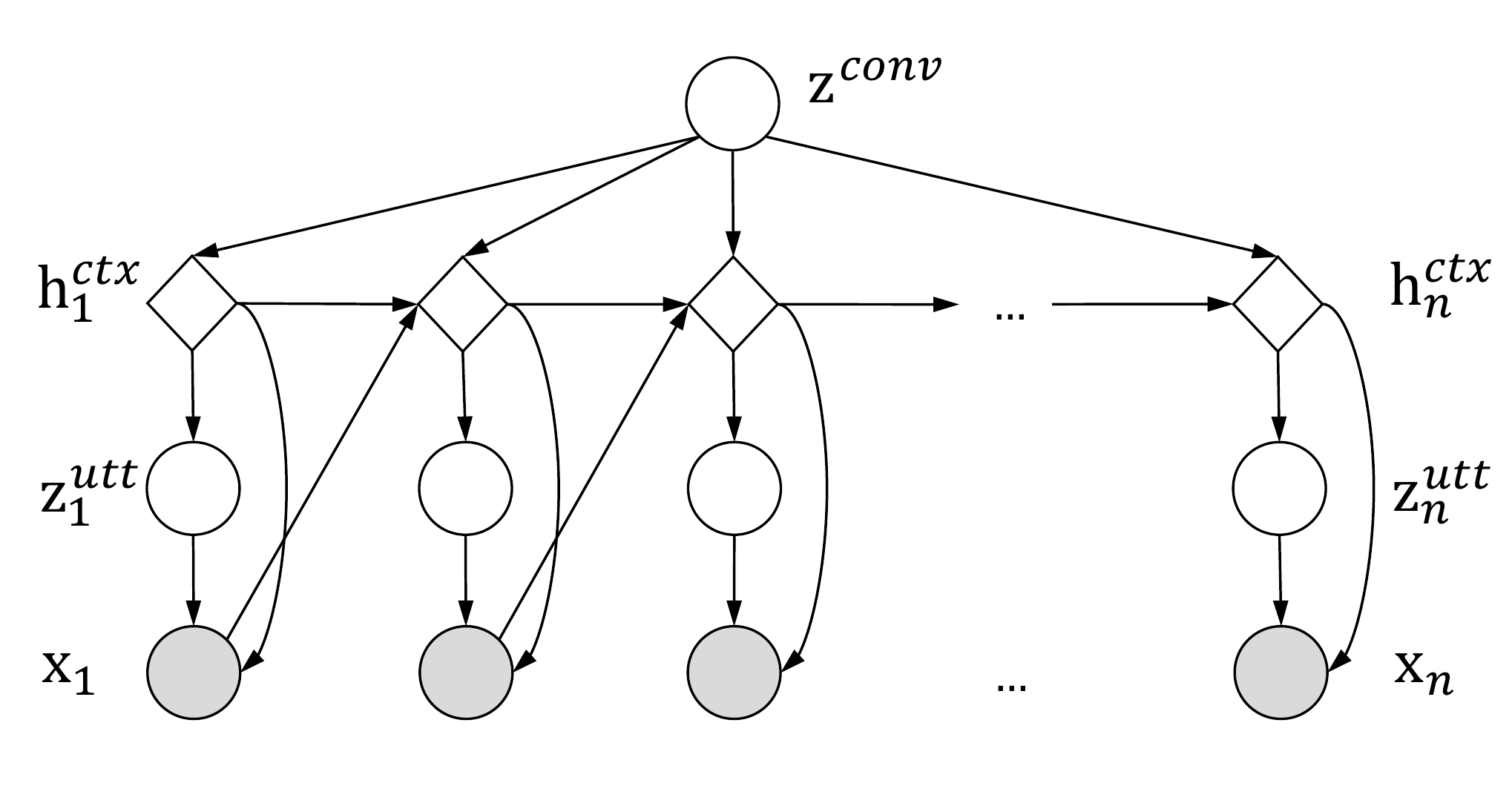}
	\caption{\textbf{Graphical representation of the Variational Hierarchical Conversation RNN (VHCR)}. The global latent variable $\vct{z}^{conv}$ provides a global context in which the conversation takes place.
	}
	\label{fig:graphical_conv}
	\vspace{-0.3cm}
\end{figure}

As discussed, we argue that the two main causes of degeneration are i) the expressiveness of the hierarchical RNN decoders, and ii) the conditional VAE structure that induces data sparsity.
This finding hints us that in order to train a non-degenerate latent variable model, we need to design a model that provides an appropriate way to regularize the hierarchical RNN decoders and alleviate data sparsity per context. 
At the same time, the model should be capable of modeling complex structure of conversation. 
Based on these insights, we propose a novel VAE structure named  Variational Hierarchical Conversation RNN (VHCR), whose graphical model is illustrated in Fig.~\ref{fig:graphical_conv}.
Below we first describe the model, and discuss its unique features.

We introduce a global conversation latent variable $\vct{z}^{\text{conv}}$ which is responsible for generating a sequence of utterances of a conversation $\vct{c}=\{\vct{x}_1, \dots, \vct{x}_n\}$:
\begin{align}
p_{\vct{\theta}}(\vct{c} | \vct{z}^{\text{conv}}) &= p_{\vct{\theta}}(\vct{x}_1, \dots, \vct{x}_n | \vct{z}^{\text{conv}})
\end{align}

Overall, the VHCR builds upon the hierarchical RNNs, following the VHRED~\citep{vhred}. 
One key update is to form a hierarchical latent structure, by using the global latent variable $\vct{z}^{\text{conv}}$ per conversation, along with local the latent variable $\vct{z}_t^{\text{utt}}$ injected at each utterance (Fig.~\ref{fig:graphical_conv}):
\begin{align}
&\vct{h}^{\text{enc}}_{t} = f^{\text{enc}}_{\vct{\theta}}(\vct{x}_{t}) \\
&\vct{h}^{\text{cxt}}_t = 
\begin{cases}
    \text{MLP}_{\vct{\theta}}(\vct{z}^{\text{conv}}), & \text{if } t=0\\
    f^{\text{cxt}}_{\vct{\theta}}(\vct{h}^{\text{cxt}}_{t-1}, \vct{h}^{\text{enc}}_{t-1}, \vct{z}^{\text{conv}}) ,              & \text{otherwise}
\end{cases} \nonumber \\
&p_{\vct{\theta}}(\vct{x}_t | \vct{x}_{<t}, \vct{z}^{\text{utt}}_t, \vct{z}^{\text{conv}}) = f^{\text{dec}}_{\vct{\theta}}(\vct{x} |\vct{h}^{\text{cxt}}_t, \vct{z}^{\text{utt}}_t, \vct{z}^{\text{conv}})  \nonumber \\
&p_{\vct{\theta}}(\vct{z}^{\text{conv}}) = \mathcal{N}(\vct{z} | \vct{0},  \vct{I}) \\ 
&p_{\vct{\theta}}(\vct{z}^{\text{utt}}_t | \vct{x}_{<t}, \vct{z}^{\text{conv}}) = \mathcal{N}(\vct{z} | \vct{\mu}_t, \vct{\sigma}_t \vct{I}) \\
&\hspace{6pt}\mbox{where } \vct{\mu}_t = \text{MLP}_{\vct{\theta}}(\vct{h}^{\text{cxt}}_t, \vct{z}^{\text{conv}}) \\
&\hspace{36pt}\vct{\sigma}_t = \text{Softplus(MLP}_{\vct{\theta}}(\vct{h}^{\text{cxt}}_t, \vct{z}^{\text{conv}})).
\end{align}

For inference of $\vct{z}^{\text{conv}}$, we use a bi-directional RNN denoted by $f^{\text{conv}}$, which runs over the utterance vectors generated by the encoder RNN:
\begin{align}
&q_{\vct{\phi}}(\vct{z}^{\text{conv}} | \vct{x}_1, ..., \vct{x}_n) = \mathcal{N}(\vct{z} | \vct{\mu}^{\text{conv}}, \vct{\sigma}^{\text{conv}} I) \label{inference} \\
&\hspace{6pt}\mbox{where } \vct{h}^{\text{conv}} = f^{\text{conv}}(\vct{h}^{\text{enc}}_1, ..., \vct{h}^{\text{enc}}_n) \label{inference_h} \\
&\hspace{36pt}\vct{\mu}^{\text{conv}} = \text{MLP}_{\vct{\phi}}(\vct{h}^{\text{conv}}) \\
&\hspace{36pt}\vct{\sigma}^{\text{conv}} = \text{Softplus}(\text{MLP}_{\vct{\phi}}(\vct{h}^{\text{conv}})).
\end{align} 
The posteriors for local variables $\vct{z}^{\text{utt}}_t$ are then conditioned on $\vct{z}^{\text{conv}}$:
\begin{align}
&q_{\vct{\phi}}(\vct{z}^{\text{utt}}_t | \vct{x}_1, ..., \vct{x}_n, \vct{z}^{\text{conv}}) = \mathcal{N}(\vct{z} | \vct{\mu}'_t, \vct{\sigma}'_t I)\\
&\hspace{6pt}\mbox{where } \vct{\mu}'_t = \text{MLP}_{\vct{\phi}}(\vct{x}_t, \vct{h}^{\text{cxt}}_t, \vct{z}^{\text{conv}}) \\
&\hspace{36pt}\vct{\sigma}'_t = \text{Softplus}(\text{MLP}_{\vct{\phi}}(\vct{x}_t, \vct{h}^{\text{cxt}}_t, \vct{z}^{\text{conv}})). \nonumber 
\end{align}

\begin{figure}[t]
	\centering
	\includegraphics[width=0.47\textwidth]{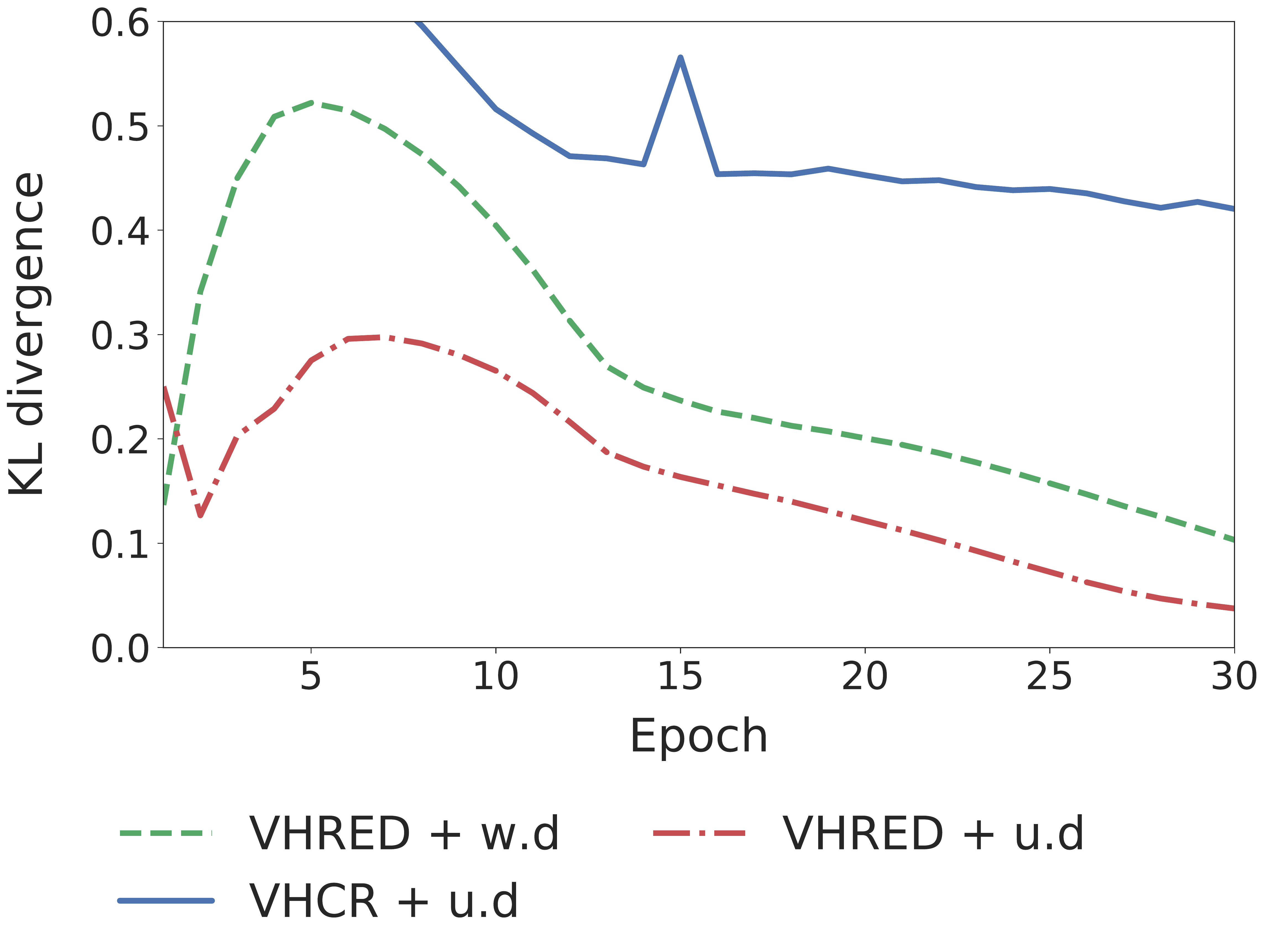}
	\vspace{-0.3cm}
	\caption{\textbf{The comparison of KL divergences}. The VHCR with the utterance drop shows high and stable KL divergence, indicating the active use of latent variables. w.d and u.d denote the word drop and the utterance drop, respectively. 
	}
	\vspace{-0.2cm}
	\label{fig:kl_compare}
\end{figure}

Our solution of VHCR to the degeneration problem is based on two ideas.
The first idea is to build a hierarchical latent structure of $\vct{z}^{\text{conv}}$ for a conversation and $\vct{z}^{\text{utt}}_t$ for each utterance. 
As $\vct{z}^{\text{conv}}$ is independent of the conditional structure, it does not suffer from the data sparsity problem. 
However, the expressive power of hierarchical RNN decoders makes the model still prone to ignore latent variables $\vct{z}^{\text{conv}}$ and $\vct{z}^{\text{utt}}_t$. 
Therefore, our second idea is to apply an \textit{utterance drop} regularization to effectively regularize the hierarchical RNNs, in order to facilitate the use of latent variables.
That is, at each time step, the utterance encoder vector $\vct{h}^{\text{enc}}_{t}$ is randomly replaced with a generic unknown vector $\vct{h}^{\text{unk}}$ with a probability $p$.  
This regularization weakens the autoregressive power of hierarchical RNNs and as well alleviates the data sparsity problem, since it induces noise into the context vector $\vct{h}^{\text{cxt}}_t$ which conditions the decoder RNN.
The difference with the word drop \citep{sentence_vae} is that our utterance drop depresses the hierarchical RNN decoders as a whole, while the word drop only weakens the lower-level decoder RNNs.
Fig.~\ref{fig:kl_compare} confirms that with the utterance drop with a probability of $0.25$, the VHCR effectively learns to use latent variables, achieving a significant degree of KL divergence. 

\subsection{Effectiveness of Hierarchical Latent Structure}
\label{sec:hiearachical}

Is the hierarchical latent structure of the VHCR crucial for effective utilization of latent variables? 
We investigate this question by applying the utterance drop on the VHRED which lacks any hierarchical latent structure. 
We observe that the KL divergence still vanishes (Fig.~\ref{fig:kl_compare}), even though the utterance drop injects considerable noise in the context $\vct{h}^{\text{cxt}}_t$. 
We argue that the utterance drop weakens the context RNN, thus it consequently fail to predict a reasonable prior distribution for $\vct{z}^{\text{utt}}$ (Eq.~\ref{vhred_prior}-\ref{vhred_prior_var}). 
If the prior is far away from the region of $\vct{z}^{\text{utt}}$ that can generate a correct target utterance, 
encoding information about the target in the variational posterior will incur a large KL divergence penalty.
If the penalty outweighs the gain of the reconstruction term in Eq.~\ref{elbo1},
then the model would learn to ignore $\vct{z}^{\text{utt}}$, in order to maximize the variational lower-bound in Eq.~\ref{elbo1}.

On the other hand, the global variable $\vct{z}^{\text{conv}}$ allows the VHCR to predict a reasonable prior for local variable $\vct{z}^{\text{utt}}_t$ even in the presence of the utterance drop regularization. 
That is, $\vct{z}^{\text{conv}}$ can act as a \textit{guide} for $\vct{z}^{\text{utt}}$ by encoding  the information for local variables. 
This reduces the KL divergence penalty induced by encoding information in $\vct{z}^{\text{utt}}$ to an affordable degree at the cost of KL divergence caused by using $\vct{z}^{\text{conv}}$. 
This trade-off is indeed a fundamental strength of hierarchical models that provide \textit{parsimonious} representation; if there exists any shared information among the local variables, it is coded in the global latent variable reducing the code length by effectively reusing the information. 
The remaining local variability is handled properly by the decoding distribution and local latent variables. 

The global variable $\vct{z}^{\text{conv}}$ provides other benefits by representing a latent global structure of a conversation, such as a topic, a length, and a tone of the conversation. 
Moreover, it allows us to control such global properties,  which is impossible for models without hierarchical latent structure.

\section{Results}
\label{section:Results}

We first describe our experimental setting, such as datasets and baselines (section \ref{section:exp_setting}). 
We then report quantitative comparisons using three different metrics (section \ref{section:results_nll}--\ref{section:results_human}).
Finally, we present qualitative analyses, including several utterance control tasks that are enabled by the hierarchal latent structure of our VHCR (section \ref{section:results_qualitative}).
We defer implementation details and additional experiment results to the appendix. 

\subsection{Experimental Setting}
\label{section:exp_setting}


\begin{table}[t]
	\caption{\textbf{Results of Negative Log-likelihood.} The inequalities denote the variational bounds. w.d and u.d., and bow denote the word drop, the utterance drop, and the auxiliary bag-of-words loss respectively. 
	}
	\label{table:result}
	\vspace{-0.2cm}
	\centering
	\small
	\subcaption{Cornell Movie Dialog}
	\vspace{-0.2cm}
	\begin{tabular}{|c|c c c|}
		\hline
		\textbf{Model} 	& NLL			& Recon. 	& KL div.	\\ 
		\hline 
		HRED			& 3.873			& -			& -		 	\\ 
		VHRED 		 	& $\le$ 3.912	& 3.619		& 0.293		\\
		VHRED + w.d		& $\le$ 3.904	& 3.553		& 0.351		\\
		VHRED + bow		& $\le$ 4.149	& 2.982		& 1.167		\\
		VHCR + u.d		& $\le$	4.026	& 3.523		& 0.503		\\ 
		\hline
	\end{tabular}	
	
	\vspace{0.2cm}
	\subcaption{Ubuntu Dialog}
	\vspace{-0.2cm}
	\begin{tabular}{|c|c c c|}	
		\hline
		\textbf{Model} 	& NLL			& Recon. 	& KL div.	\\ 
		\hline
		HRED			& 3.766			& -			& -			\\ 	
		VHRED 			& $\le$	3.767	& 3.654		& 0.113		\\
		VHRED + w.d		& $\le$ 3.824	& 3.363		& 0.461		\\ 	
		VHRED + bow		& $\le$	4.237	& 2.215		& 2.022		\\
		VHCR + u.d		& $\le$	3.951	& 3.205		& 0.756		\\ 	\hline
	\end{tabular}
	\vspace{-0.2cm}
\end{table}

\begin{table}[t]
	\caption{\textbf{KL divergence decomposition}.}
	\label{table:kl_decomposition}
	\centering
	\small
	\setlength{\tabcolsep}{5pt}
	\begin{tabular}{|c|c c c|c c c|}
		\hline
		& \multicolumn{3}{|c|}{Cornell}		& \multicolumn{3}{|c|}{Ubuntu}		\\	
		\hline
		\textbf{Model}	& Total	& $\vct{z}^{\text{conv}}$ 	& $\vct{z}^{\text{utt}}$
		& Total	& $\vct{z}^{\text{conv}}$	& $\vct{z}^{\text{utt}}$\\
		\hline
		VHRED			& 0.351	& -							& 0.351	
		& 0.461	& -							& 0.461	\\
		VHCR			& 0.503 & 0.189 					& 0.314		
		& 0.756	& 0.198 					& 0.558	\\
		\hline
	\end{tabular}
	\vspace{-0.2cm}
\end{table}

\textbf{Datasets}.
We evaluate the performance of conversation generation using two benchmark datasets: 1) Cornell Movie Dialog Corpus~\citep{movie_dialogue}, containing 220,579 conversations from 617 movies. 
2) Ubuntu Dialog Corpus \citep{ubuntu_corpus}, containing  about 1 million multi-turn conversations from Ubuntu IRC channels. 
In both datasets, we truncate utterances longer than 30 words. 

\textbf{Baselines}.
We compare our approach with four baselines. 
They are combinations of two state-of-the-art models of conversation generation with different solutions to the degeneracy.
(i) Hierarchical recurrent encoder-decoder~(HRED)~\citep{hrnn_dialogue}, 
(ii) Variational HRED~(VHRED)~\citep{vhred}, 
(iii) VHRED with the word drop~\citep{sentence_vae}, and
(iv) VHRED with the bag-of-words (bow) loss~\citep{vae_bow}.

\textbf{Performance Measures}.
Automatic evaluation of conversational systems is still a challenging problem~\citep{not_eval_dialogue}. 
Based on literature, we report three quantitative metrics: i) the negative log-likelihood (the variational bound for variational models), ii) embedding-based metrics~\cite{vhred}, and iii) human evaluation via Amazon Mechanical Turk (AMT).



\subsection{Results of Negative Log-likelihood}
\label{section:results_nll}

Table \ref{table:result} summarizes the per-word negative log-likelihood (NLL) evaluated on the test sets of two datasets.
For variational models, we instead present the variational bound of the negative log-likelihood in Eq.~\ref{elbo1},
which consists of the reconstruction error term and the KL divergence term. 
The KL divergence term can measure how much each model utilizes the latent variables. 

We observe that the NLL is the lowest by the HRED.
Variational models show higher NLLs, because they are regularized methods that are forced to rely more on latent variables. Independent of NLL values, we later show that the latent variable models often show better generalization performance in terms of embedding-based metrics and human evaluation. 
In the VHRED, the KL divergence term gradually vanishes even with the word drop regularization; thus, early stopping is necessary to obtain a meaningful KL divergence.
The VHRED with the bag-of-words loss (bow) achieves the highest KL divergence, however, at the cost of high NLL values. 
That is, the variational lower-bound minimizes the minimum description length, to which the bow loss works in an opposite direction by forcing latent variables to encode bag-of-words representation of utterances. 
Our VHCR achieves stable KL divergence without any auxiliary objective, and the NLL is lower than the VHRED + bow model.

Table \ref{table:kl_decomposition} summarizes how global and latent variable are used in the VHCR. We observe that VHCR encodes a significant amount of information in the global variable $\vct{z}^{\text{conv}}$ as well as in the local variable $\vct{z}^{\text{utt}}$, indicating that the VHCR successfully exploits its hierarchical latent structure.

\subsection{Results of Embedding-Based Metrics}
\label{section:results_embedding}


\begin{table}[t]
	\caption{\textbf{Results of embedding-based metrics.} 
		1-turn and 3-turn responses of models per context.
	}
	\label{table:embedding_metric}
	\centering
	\small
	\vspace{-0.2cm}
	\subcaption{Cornell Movie Dialog}
	\vspace{-0.2cm}	
	\begin{tabular}{|c | c c c|}
		\hline
		\textbf{Model} 	& Average	& Extrema 	& Greedy \\	
		\hline	
		\multicolumn{4}{|c|}{1-turn} \\ \hline
		HRED			& 0.541		& 0.370		& 0.387	\\ 	
		VHRED 			& 0.543		& 0.356		& 0.393	\\
		VHRED + w.d		& 0.554		& 0.365		& 0.404	\\ 	
		VHRED + bow		& 0.555		& 0.350		& 0.411	\\
		VHCR + u.d		& \textbf{0.585}&\textbf{0.376}&\textbf{0.434} \\
		\hline
		\multicolumn{4}{|c|}{3-turn} 		\\ 
		\hline
		HRED			& 0.556		& 0.372		& 0.395	\\ 	
		VHRED 			& 0.554		& 0.360		& 0.398		\\
		VHRED + w.d		& 0.566		& 0.369		& 0.408 \\ 	
		VHRED + bow		& 0.573		& 0.360		& 0.423	\\ 
		VHCR + u.d		& \textbf{0.588}&\textbf{0.378}&\textbf{0.429}\\ 
		\hline
	\end{tabular}
	
	\vspace{0.2cm}
	\subcaption{Ubuntu Dialog}
	\vspace{-0.2cm}
	\begin{tabular}{|c | c c c|}
		\hline
		\textbf{Model} 	& Average	& Extrema 	& Greedy	\\	
		\hline	
		\multicolumn{4}{|c|}{1-turn} \\ \hline
		HRED			& 0.567		&\textbf{0.337}& 0.412	\\ 	
		VHRED 			& 0.547		& 0.322		& 0.398 \\
		VHRED + w.d		& 0.545 	& 0.314		& 0.398	\\ 	
		VHRED + bow		& 0.545		& 0.306		& 0.398	\\
		VHCR + u.d		&\textbf{0.570}& 0.312	& \textbf{0.425}	\\
		\hline
		\multicolumn{4}{|c|}{3-turn} \\ \hline
		HRED			& 0.559		&\textbf{0.324}& 0.402	\\ 	
		VHRED 			& 0.551		& 0.315		& 0.397	\\
		VHRED + w.d		& 0.551		& 0.309		& 0.399 \\ 	
		VHRED + bow		& 0.552		& 0.303		& 0.398\\ 
		VHCR + u.d		&\textbf{0.574}& 0.311	& \textbf{0.422}	\\ 	
		\hline
	\end{tabular}
	\vspace{-0.2cm}
\end{table}

\begin{table*}[t]
	\caption{\textbf{Results of human evaluation via AMT}. Human turkers are asked to choose which response is more appropriate in a given context, without knowing which algorithms generate which responses. For each pair of models, we carry out three evaluation batches, each of which consists of 100 random test samples evaluated by five unique humans. We report mean preferences with $\pm 90\%$ confidence interval.}
	\label{table:human_eval}
	\centering
	\small
	\vspace{-0.2cm}	
	\begin{tabular}{|c|c c c| c c c|}
		\hline
		&	& Cornell & & & Ubuntu & \\
		\hline
		Opponent 			& Wins 						& Losses 		& Ties 			& Wins 		& Losses 	& Ties \\
		\hline
		VHCR vs HRED 		& $\mathbf{28.5 \pm 1.9}$	& $28.2 \pm 1.9$& $43.3 \pm 2.1$
		& $\mathbf{52.9 \pm 2.1}$	& $42.2 \pm 2.1$& $4.9 \pm 0.9$ 	\\
		VHCR vs VHRED + w.d & $\mathbf{29.9 \pm 1.9}$	& $28.0 \pm 1.9$& $42.1 \pm 2.1$
		& $\mathbf{48.1 \pm 2.1}$	& $40.1 \pm 3.6$& $11.9 \pm 1.4$ 	\\
		VHCR vs VHRED + bow	& $\mathbf{31.3 \pm 2.0}$	& $26.9 \pm 1.9$& $41.7 \pm 2.1$
		& $\mathbf{46.1 \pm 2.1}$	& $39.9 \pm 2.1$& $14.0 \pm 1.5$ \\
		\hline
	\end{tabular}	
	
	\vspace{-0.1cm}		
\end{table*}

\begin{table*}[t]
	\caption{\textbf{Qualitative comparison of generated responses}. Top two rows show the samples from Cornell Movie Dialog, while the bottom two rows are from Ubuntu Dialog.}
	\label{table:sample_comparison}
	\centering
	\scriptsize
	\vspace{-0.2cm}
	\begin{tabular}{|p{0.30\linewidth}|p{0.12\linewidth}|p{0.15\linewidth}|p{0.15\linewidth}|p{0.15\linewidth}|}
		\hline                                                                      
		Context	& HRED & VHRED & VHRED+bow & VHCR \\
		\hline
		i'm going home. $\rightarrow$
		what's wrong? $\rightarrow$
		i don't know. one of these days, i'm gon na see louis. my back's killing me. 
		& i don't know.
		& i don't want to be here. i'm going to be here.
		& are you going to go now?
		& this isn't a joke. you're going to be here. \\
		\hline
		how's sleeping beauty? $\rightarrow$
		still sleeping. $\rightarrow$
		kiss him, maybe he'll turn into a frog.
		& he's a good man.
		& what?
		& we won't?
		& maybe he's dead. \\
		\hline
		when i mount my usb drive i get a permission denied but i can easily see the files if i boot via live cd any ideas on how i $\rightarrow$
		the file permissions of the ext3 file system are likely preventing you $\rightarrow$
		thanks, any idea on options to fix?
		& no, i don't know how to fix it, but i'm not sure how to fix it
		& what kind of filesystem is it on?
		& you did it just first? then you'll need to reboot and then remount it up again, if your problem persists, i'd give you the
		& the only thing i can think of is to fix the permissions issue, try mounting the drive with the uid and gid options in the forum \\
		\hline
		hello i need some support, anybody there? $\rightarrow$
		go ahead with your questions. $<$unk$>$ are not expected on irc. $\rightarrow$
		i have an hp pavilion $<$unk$>$ here and i can't get it to install ubuntu
		& what's the problem?
		& $<$unk$>$. what's the problem?
		& i'm sorry.
		& they should be supported by canonical. you don't need to use them. \\
		\hline
	\end{tabular}
	\vspace{-0.4cm}
\end{table*}

\begin{table}[t]
	\caption{\textbf{An example of interpolated 3-turn responses over $\vct{z}^{\text{conv}}$ on Cornell Movie Dialog}.}
	\label{table:interpolate_cornell}
	\centering
	\scriptsize
	\vspace{-0.2cm}
	\begin{tabular}{|p{0.95\linewidth}|}
		\hline
		\textbf{nice to meet you.}                                 \\                                                       
		$\rightarrow$ \textbf{where's my wife?}                                 \\                                                       
		$\rightarrow$ \textbf{she's asleep.}                                    \\                                                                                                                                 
		\hline
		go on, frank.                                    \\                                                       
		$\rightarrow$ i ll tell you what i'm doing. i'm a little tired. \\
		$\rightarrow$ why?                                              \\                                                       
		\hline                                                                                  
		now, you're not going.                          \\                                                       
		$\rightarrow$ not for a long time.                              \\                                                       
		$\rightarrow$ but you're a little $<$unk$>$.                       \\                                                       
		\hline
		get out of here.                                  \\                                                       
		$\rightarrow$ all right.                                        \\                                                       
		$\rightarrow$ you want to go home?                              \\                                                       
		\hline
		\textbf{get out of here!!!}                              \\                                                       
		$\rightarrow$ \textbf{it's not your fault, is n't it?}                 \\                                                       
		$\rightarrow$ \textbf{why? what's wrong?}      						   \\
		\hline                           
	\end{tabular}
	\vspace{-0.3cm}
\end{table}

\begin{table}[t]
	\caption{\textbf{An example of 3-turn responses conditioned on sampled $\vct{z}^{\text{utt}}$ for a single fixed $\vct{z}^{\text{conv}}$}.}
	\label{table:conditioned_cornell}
	\centering
	\scriptsize
	\vspace{-0.2cm}
	\begin{tabular}{|p{0.95\linewidth}|}
		\hline
		where is she? \\                                                                                                        
		$\rightarrow$ she's the only one who knows where she is, she's going to be all right. \\                                           
		$\rightarrow$ oh, you're the only one who's gon na be. she's a $<$unk$>$.   \\                                                       
		\hline
		where's my wife?                                \\                                                                     
		$\rightarrow$ you've got to get out of here, you know? you're the one who's gon na be here.             \\                      
		$\rightarrow$ oh, that's nice.             \\                                                                                       
		\hline
		where are you?                                                             \\                                           
		$\rightarrow$ well, i was just thinking about you and i know what you're doing. i'm going to have to go to the $<$unk$>$ and i'm    \\
		$\rightarrow$ i'm sorry.                             \\                                                                     
		\hline
		where are you going?        \\                                                                                          
		$\rightarrow$ to get you to the airport.    \\                                                                                        
		$\rightarrow$ you're going to be late?     \\                                                                    
		\hline
		where are you going?                                         \\                                                         
		$\rightarrow$ to the $<$unk$>$. i am not going to tell you what i am. i am the only one who has to be. i will be the    \\              
		$\rightarrow$ you've got to stop!    \\                                                                                
		\hline                           
	\end{tabular}
	\vspace{-0.3cm}
\end{table}

The embedding-based metrics \citep{vhred, embedding_metric} measure the textual similarity between the words in the model response and the ground truth. 
We represent words using Word2Vec embeddings trained on the Google News Corpus\footnote{https://code.google.com/archive/p/word2vec/.}. 
The \textit{average} metric projects each utterance to a vector by taking the mean over word embeddings in the utterance, 
and computes the cosine similarity between the model response vector and the ground truth vector. 
The \textit{extrema} metric is similar to the average metric, only except that it takes the extremum of each dimension, instead of the mean.
The \textit{greedy} metric first finds the best non-exclusive word alignment between the model response and the ground truth, 
and then computes the mean over the cosine similarity between the aligned words.  

Table ~\ref{table:embedding_metric} compares the different methods with three embedding-based metrics.
Each model generates a single response (1-turn) or consecutive three responses (3-turn) for a given context.
For 3-turn cases, we report the average of metrics measured for three turns. 
We use the greedy decoding for all the models.  

Our VHCR achieves the best results in most metrics.
The HRED is the worst on the Cornell Movie dataset, but outperforms the VHRED and VHRED + bow on the Ubuntu Dialog dataset. 
Although the VHRED + bow shows the highest KL divergence, its performance is similar to that of VHRED, and worse than that of the VHCR model. 
It suggests that a higher KL divergence does not necessarily lead to better performance; it is more important for the models to balance the modeling powers of the decoder and the latent variables.
The VHCR uses a more sophisticated hierarchical latent structure, which better reflects the structure of natural language conversations.  



\subsection{Results of Human Evaluation}
\label{section:results_human}

Table~\ref{table:human_eval} reports human evaluation results via Amazon Mechanical Turk (AMT). 
The VHCR outperforms the baselines in both datasets; yet the performance improvement in Cornell Movie Dialog are less significant compared to that of Ubuntu.
We empirically find that Cornell Movie dataset is small in size, but very diverse and complex in content and style, 
and the models often fail to generate sensible responses for the context. 
The performance gap with the HRED is the smallest, suggesting that the VAE models without hierarchical latent structure have overfitted to Cornell Movie dataset.

\subsection{Qualitative Analyses}
\label{section:results_qualitative}

\textbf{Comparison of Predicted Responses}.
Table~\ref{table:sample_comparison} compares the generated responses of algorithms. 
Overall, the VHCR creates more consistent responses within the context of a given conversation. 
This is supposedly due to the global latent variable $\vct{z}^{\text{conv}}$ that provides a more direct and effective way to handle the global context of a conversation.
The context RNN of the baseline models can handle long-term context to some extent, but not as much as the VHCR. 

\textbf{Interpolation on $\vct{z}^{\text{conv}}$}.
We present examples of one advantage by the hierarchical latent structure of the VHCR, which cannot be done by the other existing models.
Table~\ref{table:interpolate_cornell} shows how the generated responses vary according to the interpolation on $\vct{z}^{\text{conv}}$. 
We randomly sample two $\vct{z}^{\text{conv}}$ from a standard Gaussian prior as references (\ie the top and the bottom row of Table~\ref{table:interpolate_cornell}),
and interpolate points between them. 
We generate 3-turn conversations conditioned on given $\vct{z}^{\text{conv}}$.
We see that $\vct{z}^{\text{conv}}$ controls the overall tone and content of conversations;
for example, the tone of the response is friendly in the first sample, but gradually becomes hostile as $\vct{z}^{\text{conv}}$ changes.  

\textbf{Generation on a Fixed $\vct{z}^{\text{conv}}$}.
We also study how fixing a global conversation latent variable $\vct{z}^{\text{conv}}$ affects the conversation generation. 
Table~\ref{table:conditioned_cornell} shows an example, where 
we randomly fix a reference $\vct{z}^{\text{conv}}$ from the prior, and generate multiple examples of 3-turn conversation using randomly sampled local variables $\vct{z}^{\text{utt}}$.
We observe that $\vct{z}^{\text{conv}}$ heavily affects the form of the first utterance; in the examples, the first utterances all start with a ``where'' phrase. 
At the same time, responses show variations according to different local variables $\vct{z}^{\text{utt}}$. 
These examples show that the hierarchical latent structure of VHCR allows both global and fine-grained control over generated conversations.

\section{Discussion}

We introduced the variational hierarchical conversation RNN (VHCR) for conversation modeling. We noted that the degeneration problem in existing VAE models such as the VHRED is persistent, and proposed a hierarchical latent variable model with the utterance drop regularization.
Our VHCR obtained higher and more stable KL divergences than various versions of VHRED models without using any auxiliary objective.
The empirical results showed that the VHCR better reflected the structure of natural conversations, and outperformed previous models.
Moreover, the hierarchical latent structure allowed both global and fine-grained control over the conversation generation.

%
%

\section*{Acknowledgments}
This work was supported by Kakao and Kakao Brain corporations, and Creative-Pioneering Researchers Program through Seoul National University. 
Gunhee Kim is the corresponding author. 

\bibliography{refs}
\bibliographystyle{acl_natbib}

\appendix
\section{Data Processing}

In both datasets, we truncate utterances longer than 30 words. Tokenization and text preprocessing is carried out using Spacy \footnote{https://spacy.io/}.  

As Cornell Movie Dialog does not provide a separate test set, we randomly choose 80\% of the conversations in Cornell Movie Dialog as training set. The remaining 20\% is evenly split into validation set and test set.

\section{Implementation Details}

We use Pytorch Framework \footnote{http://pytorch.org/} for our implementations. We plan to release our code public.

We build a dictionary with the vocabulary size of 20,000, and further remove words with frequency less than five. 
We set the word embedding dimension to 500. 
We adopt Gated Recurrent Unit (GRU)~\citep{rnn_phrase_representation} in our model and all baseline models, 
as we observe no improvement of LSTMs~\citep{lstm} over GRUs in our experiments. 
We use one-layer GRU with the hidden dimension of 1,000 (2,000 for bi-directional GRU) for our RNN decoders. 
Two-layer MLPs with hidden layer size 1000 parameterizes the distribution of latent variables. 
All latent variables have a dimension of 100. 
We apply dropout ratio of 0.2 during training. Batch size is 80 for Cornell Movie Dialog, and 40 for Ubuntu Dialog.
For optimization, we use Adam~\citep{adam} with a learning rate of 0.0001 with gradient clipping. 
We adopt early stopping by monitoring the performance on the validation set. 
We apply the KL annealing to all variational models, where the KL multiplier $\lambda$  gradually increases from 0 to 1 over 15,000 steps on Cornell Movie Dialog and over 250,000 steps on Ubuntu Dialog. For both the word drop and the utterance drop, we use drop probability of 0.25.

\section{Experimental Results}

\begin{table}[t]
	\caption{\textbf{An example of interpolated 3-turn responses over $\vct{z}^{\text{conv}}$ on Cornell Movie Dialog}.}
	\label{table:interpolate_supple}
	\centering
	\scriptsize
	\vspace{-0.2cm}
	\begin{tabular}{|p{0.95\linewidth}|}
		\hline
		\textbf{what?}                                               \\                                                                 
		$\rightarrow$ \textbf{you're a good man, you're a liar. you're a good man}.            \\                                                
		$\rightarrow$ \textbf{what's wrong with you}?                                \\                                                                                                                                                       
		\hline
		what's wrong?                                                   \\                                                    
		$\rightarrow$ it's a good idea.                                               \\                                                    
		$\rightarrow$ you're a good man, you don't have to do anything like that.    \\                                                                                                                                       
		\hline
		isn't there a problem with you?                                  \\                                                   
		$\rightarrow$ well, maybe ...                                                   \\                                                   
		$\rightarrow$ then, you'll be fine.                                           \\                                                                                                                                                           
		\hline
		isn't that a little joke, isn't it?                            \\                                                   
		$\rightarrow$ yes.                                                              \\                                                   
		$\rightarrow$ you don't have to be a hero, do you?                            \\                                                                                                
		\hline
		\textbf{isn't that a fact that you're a $<$unk$>$?}                         \\                                                    
		$\rightarrow$ \textbf{no, i'm fine.}                                                 \\                                                    
		$\rightarrow$ \textbf{then why are you here?}                                           \\                                                                                                                                      
		\hline                           
	\end{tabular}
	\vspace{-0.3cm}
\end{table}

\begin{table}[t]
	\caption{\textbf{An example of 3-turn responses conditioned on sampled $\vct{z}^{\text{utt}}$ for a single fixed $\vct{z}^{\text{conv}}$}.}
	\label{table:conditioned_supple}
	\centering
	\scriptsize
	\vspace{-0.2cm}
	\begin{tabular}{|p{0.95\linewidth}|}
		\hline                                   	
		you're a liar?                                                         \\            
		$\rightarrow$ i don't know.                                                           \\            
		$\rightarrow$ i'll tell you what. i'll be there. i'll be back in a minute.        \\                                                
		\hline
		you don't know?                                                               \\      
		$\rightarrow$ no, i mean, i don't know, i just don't think i'd be in love with you.    \\      
		$\rightarrow$ well?                                                                          \\      
		\hline
		you want a drink?                                                               \\     
		$\rightarrow$ sure.                                                                           \\     
		$\rightarrow$ good. good night, mary.                                                       \\     
		\hline
		you're sure?                                                                  \\      
		$\rightarrow$ we're going to have to do something, don't we?                              \\     
		$\rightarrow$ well, i don't know what to say.                                              \\                                          
		\hline
		you know that?                                                                 \\      
		$\rightarrow$ sure.                                                                          \\      
		$\rightarrow$ you're a good man, are n't you?                                              \\                  
		\hline                           
	\end{tabular}
	\vspace{-0.3cm}
\end{table}

\begin{table*}[t]
	\caption{\textbf{Comparison of generated responses on Cornell Movie Dialog}.}
	\label{table:sample_comparison_cornell}
	\centering
	\scriptsize
	\begin{tabular}{|p{0.30\linewidth}|p{0.12\linewidth}|p{0.15\linewidth}|p{0.15\linewidth}|p{0.15\linewidth}|}
		\hline                                                                      
		Context	& HRED & VHRED & VHRED+bow & VHCR \\
		\hline
		thanks, jack. i'll be in touch. $\rightarrow$
		do you know what you're doing. $\rightarrow$
		trust me. 
		& i'm not going to be a good man. 
		& you're not going to be a good time. 
		& you ever seen a $<$unk$>$ or something? you never seen a $<$unk$>$? 
		& how much do you know about the $<$unk$>$? \\
		\hline
		i saw desert sun, i wanna tell you ... $\rightarrow$
		no, i was, i was, i was just u learning u on, it's a ... $\rightarrow$
		how'd you like working with richard hill?
		& oh, i don't know ...
		& what did you say?
		& i don't know ... i mean, i know ... i mean, you look like this ... it's ... you know, i've been
		& i've been thinking about it, but i ... \\                     
		\hline 
		i don't think the people who live around here would like that very much ... $\rightarrow$
		do you think they like the fact that $<$unk$>$ has dropped ninety eight percent in the last ten years? $\rightarrow$
		no, but ... 
		& you're a good man, you know.
		& ... and then it's not a $<$unk$>$. it's a $<$unk$>$ of $<$unk$>$. it's a long time. it's a long time. 
		& i know. it's a beautiful woman to do with your own. so, you're not married.
		& they're not going to kill me. \\
		\hline
		every one of these women thinks she should be the next angel of the month. it's your job, ted, to decide which of them is $\rightarrow$
		let me get this straight ... $\rightarrow$
		if it isn't straight by now, son, you probably should see a doctor.
		& no, no, no, no, no, no, no, no, no, no, no, no, no, no, no,
		& ... i'm not a $<$unk$>$, i'm a $<$unk$>$ ...
		& you look dead.
		& it's not the same thing. it's not the truth. \\
		\hline
		like following me around and $<$unk$>$ the people i work with wasn't bad enough, but breaking into my apartment  $\rightarrow$
		it was open. $\rightarrow$
		you got a lotta nerve.
		& i'm not a $<$unk$>$.
		& i was a $<$unk$>$ $<$unk$>$.
		& you're not a cop. i didn't tell you. it was just a little $<$unk$>$, but it wasn't me. you're just trying to
		& why don't you just go to the police? \\
		\hline
	\end{tabular}
\end{table*}

\begin{table*}[t]
	\caption{\textbf{Comparison of generated responses on Ubuntu Dialog}.}
	\label{table:sample_comparison_ubuntu}
	\centering
	\scriptsize
	\begin{tabular}{|p{0.30\linewidth}|p{0.14\linewidth}|p{0.14\linewidth}|p{0.14\linewidth}|p{0.14\linewidth}|}
		\hline
		Context	& HRED & VHRED & VHRED+bow & VHCR \\
		\hline
		hi, everyone. i download the $<$unk$>$, and want to install it with sudo dpkg i google earth stable current i386.deb, it tells me google earth $\rightarrow$
		the error suggests running sudo apt get f install $\rightarrow$
		i tried that, it comes the same error
		& then you should be able to get the .deb from the deb file
		& that's a good idea, sorry
		& what is it, please.
		& it says error dependency is not satisfiable $<$unk$>$ that's the error i get is that \\
		\hline
		click the database you $<$unk$>$ to import it into, and look for the import button. import as sql. $\rightarrow$
		noticed. this will take me a while because i am not used to this subject and not fluent in it. $\rightarrow$
		you may need to destroy the tables in the database before importing.
		& i am not sure what you mean. i am using the default ubuntu install.
		& it is a $<$unk$>$ of the $<$unk$>$. i have a $<$unk$>$ $<$unk$>$ and i want to be able to use the database on the $<$unk$>$. i am
		& i dont think that s what you are saying is it isnt an option
		& then how to do that? \\
		\hline
		i'm trying to setup a shortcut dir of this path home user downloads in this path jail $<$unk$>$ site how do i do this? i tried using $\rightarrow$
		ln s home user downloads jail $<$unk$>$ site downloads if you're trying to do something with a chroot jail guessing from the path you can't symlink outside $\rightarrow$
		thank you for the answer. but making a vlan for each user and then limit the vlan is not possible when you only have ssh root access right
		& i'm not sure if that's possible, but i'm not sure if it's possible to do that.
		& $<$unk$>$ is the user that owns the home directory
		& i'd say that's the case ... i've never even used a shell to draw the data with the destination folder
		& look at the permissions of the folder and the other is $<$unk$>$, yes \\
		\hline
		hi i am facing a problem! while upgrading my ubuntu to 12.04 lts unfortunately shutdown the system and after that when i reboot my system it shows black
		$\rightarrow$ did the shutdown occur in the middle of the upgrade?
		$\rightarrow$ yes, one of my friend just shut down, he was not aware that upgrade is going on now the system freezes at seemingly random points and leaves
		& i 'm not sure what you mean by 'shutdown'.
		& not sure what the problem is, try sudo apt get update sudo apt get upgrade sudo apt get upgrade sudo apt get upgrade
		& it has been interrupted.
		& so you are saying that it is not able to boot the cd or something? \\
		\hline
	\end{tabular}
\end{table*}

Table \ref{table:interpolate_supple} -- \ref{table:sample_comparison_ubuntu} shows additional sample generation results. 

\section{Human Evaluation}
We perform human evaluation study on Amazon Mechanical Turk (AMT). 
We first filter out contexts that contain generic unknown word (unk) token from the test set. Using these contexts, we generate model response samples. Samples that contain less than 4 tokens are removed. The order of the samples and the order of model responses are randomly shuffled. 

Evaluation procedure is as follows: given a context and two model responses, a Turker decides which response is more appropriate in the given context. In the case where the Turker thinks that two responses are about equally good or bad or does not understand the context, we ask the Turker to choose ``tie''. 
We randomly select 100 samples to build a batch for a human intelligence test (HIT). For each pair of models, we perform 3 HITs on AMT and each HIT is evaluated by 5 unique humans. In total we obtain 9000 preferences in 90 HITs.

\end{document}